\documentclass[conference]{IEEEtran}
\IEEEoverridecommandlockouts
\usepackage{cite}
\usepackage{amsmath,amssymb,amsfonts}
\usepackage{algorithmic}
\usepackage{graphicx}
\usepackage{textcomp}
\usepackage{xcolor}

\usepackage{times}
\usepackage{graphicx}
\usepackage{wrapfig}
\usepackage{amsmath}
\usepackage{multirow}
\usepackage{booktabs}
\usepackage[labelformat=simple]{subcaption}
\usepackage{enumitem}
\usepackage{makecell}
\usepackage[hidelinks]{hyperref}

\newcommand\blfootnote[1]{%
  \begingroup
  \renewcommand\thefootnote{}\footnote{#1}%
  \addtocounter{footnote}{-1}%
  \endgroup
}

\def\BibTeX{{\rm B\kern-.05em{\sc i\kern-.025em b}\kern-.08em
    T\kern-.1667em\lower.7ex\hbox{E}\kern-.125emX}}
\begin{document}

\title{A Multi-level Alignment Training Scheme for Video-and-Language Grounding}

\author{\IEEEauthorblockN{Yubo Zhang}
\IEEEauthorblockA{\textit{Department of Computer Science} \\
\textit{ University of North Carolina at Chapel Hill}\\
Chapel Hill, NC, USA \\
zhangyb@cs.unc.edu}
\and
\IEEEauthorblockN{Feiyang Niu}
\IEEEauthorblockA{Amazon Alexa AI\\
Sunnyvale, CA, USA \\
nfeiyan@amazon.com}
\and
\IEEEauthorblockN{Qing Ping}
\IEEEauthorblockA{Amazon Alexa AI\\
Sunnyvale, CA, USA \\
pingqing@amazon.com}
\and
\IEEEauthorblockN{Govind Thattai}
\IEEEauthorblockA{Amazon Alexa AI\\
Sunnyvale, CA, USA \\
thattg@amazon.com}
}

\thispagestyle{plain}
\pagestyle{plain}

\maketitle

\begin{abstract}
To solve video-and-language grounding tasks, the key is for the network to understand the connection between the two modalities. For a pair of video and language description, their semantic relation is reflected by their encodings' similarity. A good multi-modality encoder should be able to well capture both inputs' semantics and encode them in the shared feature space where embedding distance gets properly translated into their semantic similarity. In this work, we focused on this semantic connection between video and language, and developed a multi-level alignment training scheme to directly shape the encoding process. Global and segment levels of video-language alignment pairs were designed, based on the information similarity ranging from high-level context to fine-grained semantics. The contrastive loss was used to contrast the encodings' similarities between the positive and negative alignment pairs, and to ensure the network is trained in such a way that similar information is encoded closely in the shared feature space while information of different semantics is kept apart. Our multi-level alignment training can be applied to various video-and-language grounding tasks. Together with the task-specific training loss, our framework achieved comparable performance as previous state-of-the-arts on multiple video QA and retrieval datasets. \blfootnote{Copyright 2022 IEEE International Conference on Data Mining Workshops (ICDMW): \url{https://ieeexplore.ieee.org/document/10031067}. All rights reserved.}
\end{abstract}


\section{Introduction}

Having a machine that can follow human instructions and adapt to visual surroundings is the key to building a intelligent system to aid human beings in our daily activities.
To achieve this goal, the smart system will need to understand the meaning of the input natural language, and to be aware of the information embedded in the visual input such as videos. More importantly, the system has to have the ability to make connections between the two modalities to further reason with the joint-modality information.

\begin{figure*}[htp]
  \centering
  \begin{subfigure}{\linewidth}
    \centering
    \includegraphics[width=.8\linewidth]{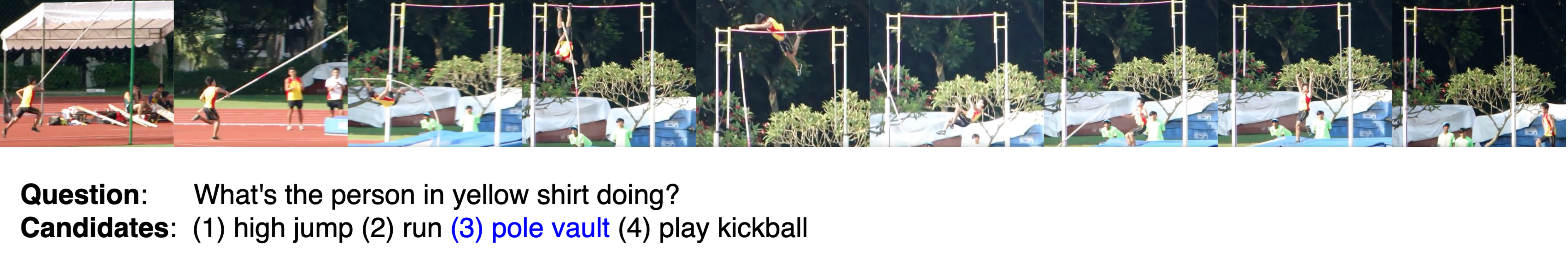}
    \vspace{-3pt}
    \caption{Video QA}
    \label{fig:video_qa}
  \end{subfigure}

  \vspace{3pt}
  \begin{subfigure}{\linewidth}
    \centering
    \includegraphics[width=.8\linewidth]{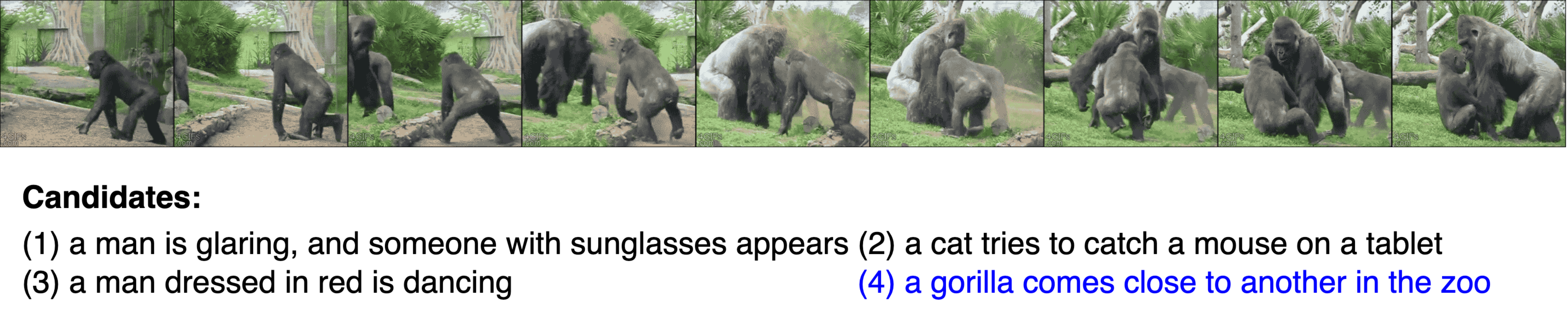}
    \vspace{-4pt}
    \caption{Video-text retrieval}
    \label{fig:video_to_text}
  \end{subfigure}  
  
  \vspace{3pt}
  \begin{subfigure}{\linewidth}
    \centering
    \includegraphics[width=.8\linewidth]{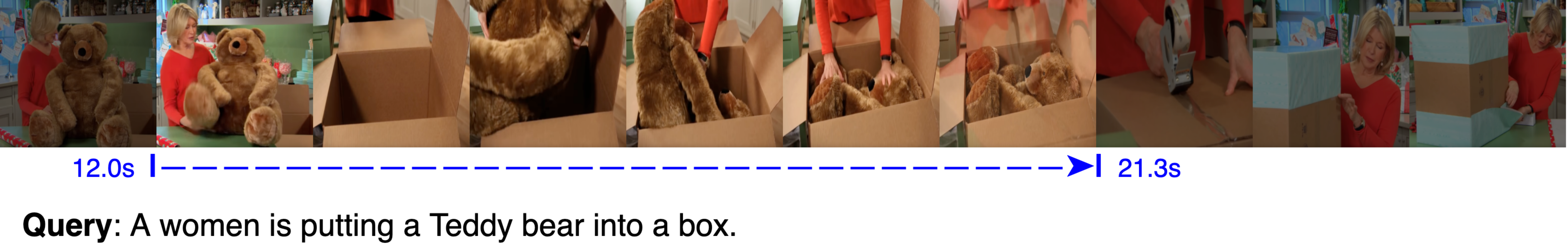}
    \vspace{-3pt}
    \caption{Video moment retrieval}
    \label{fig:video_localization}
  \end{subfigure}  
  
  \caption{Video-and-language tasks. (a) Video QA task aims to predict answers to natural language questions given a video as context. (b) Video-text retrieval selects the text description from the candidate pool that best matches the video content (vice versa, text-video retrieval). (c) Video moment retrieval localizes the video segment that aligns with the query text input. \textcolor{blue}{Ground-truths} are colored in blue.}
  \label{fig:video_language_tasks}
  
\end{figure*}

Neural network models have been proved to be powerful for understanding complex real-world information, and have shown their strength in solving video-and-language grounding problems, such as video question answering (QA) and video retrieval. Examples of three common video-and-language grounding tasks are included in Figure~\ref{fig:video_language_tasks}.
Task-specific frameworks have been developed, targeting each grounding task's own challenges.
Some auxiliary techniques, e.g., object detection and scene graph reasoning, have been explored in order to achieve better performance~\cite{xu2019scene, lee2019visual, liang2020lrta, chen2020uniter, zhu2020actbert}. 
However, these tailored models are usually lacking generalizability,
and serious modifications are often required for them to handle other types of grounding tasks~\cite{chen2020fine,le2020hierarchical}.

\begin{figure}[t]
\centering

  \begin{subfigure}[b]{0.16\textwidth}
    \includegraphics[width=\linewidth]{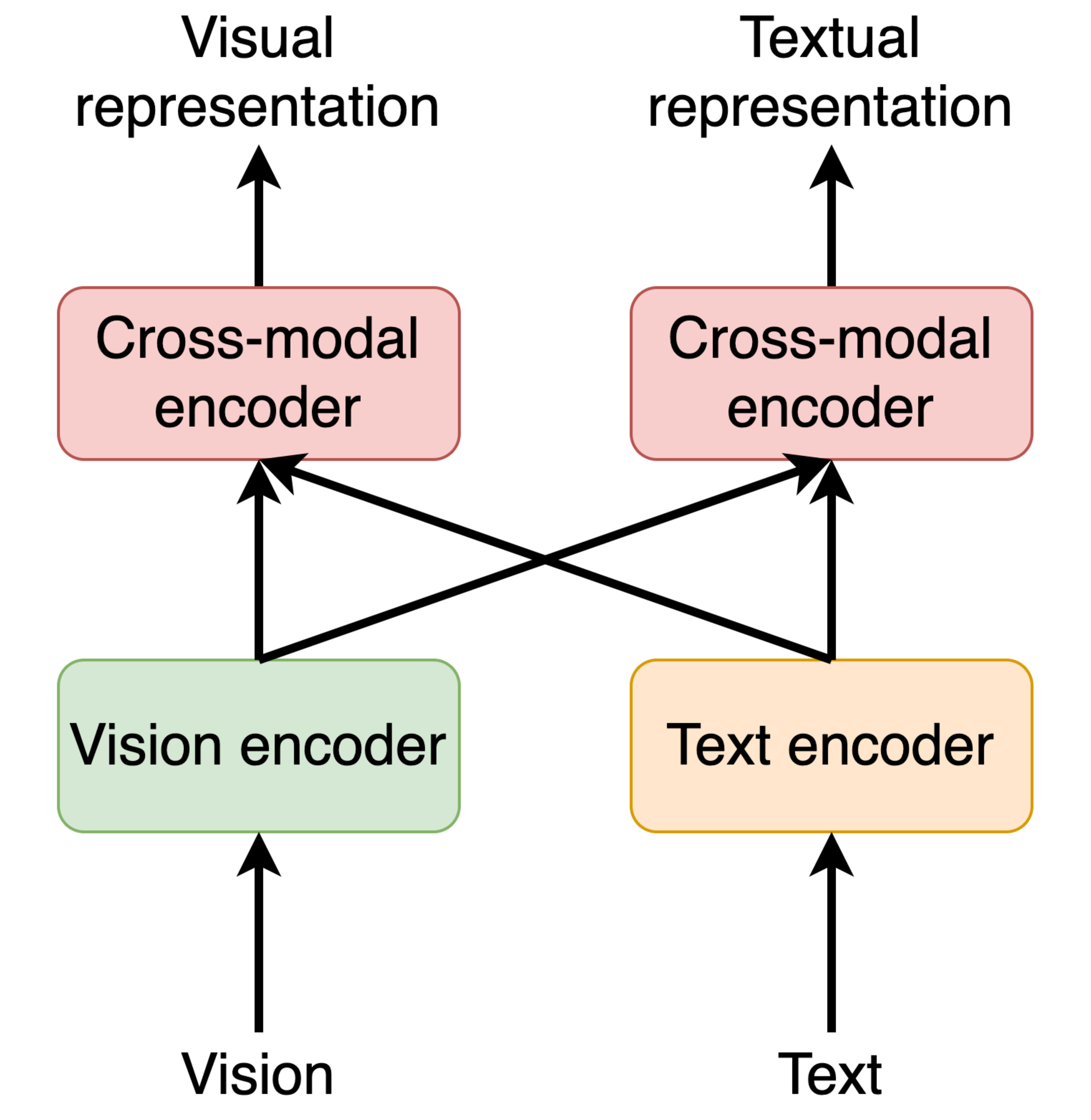}
    \caption{}
    \label{fig:cross_modality}
  \end{subfigure}%
  \hspace*{\fill}   
  \begin{subfigure}[b]{0.10\textwidth}
    \includegraphics[width=\linewidth]{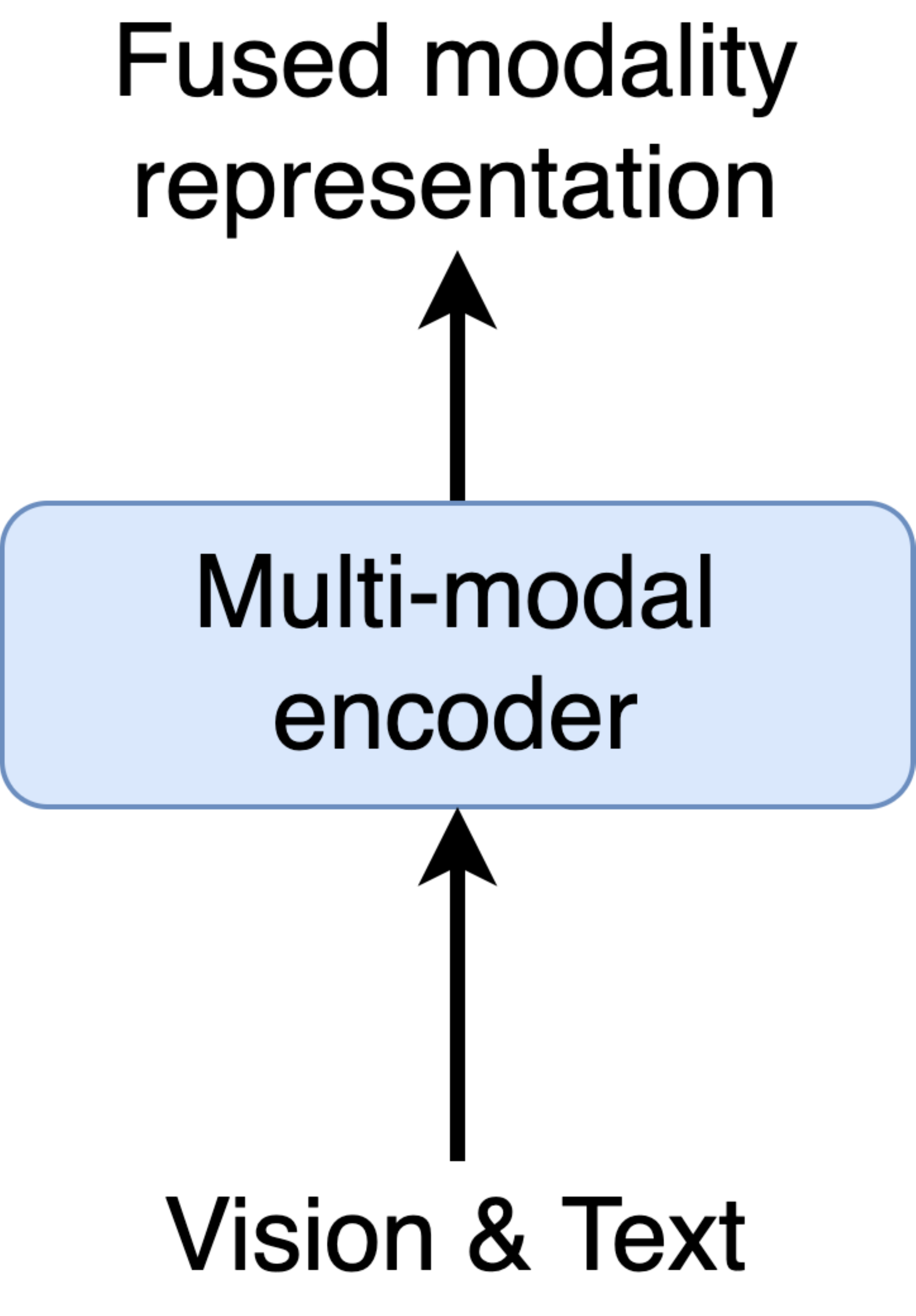}
    \caption{}
    \label{fig:joint_modality}
  \end{subfigure}%
  \hspace*{\fill}   
  \begin{subfigure}[b]{0.22\textwidth}
    \includegraphics[width=\linewidth]{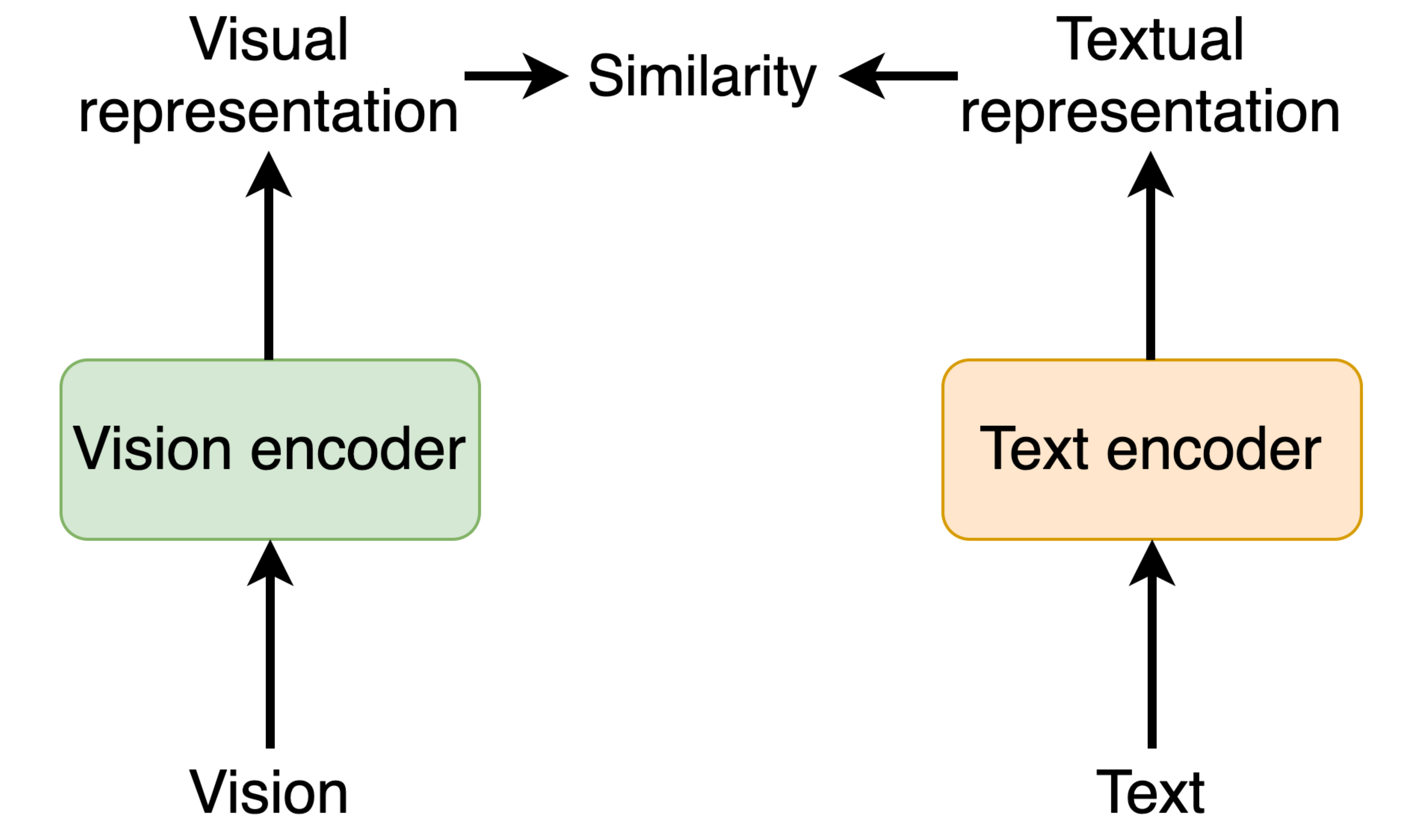}
    \caption{}
    \label{fig:separate_modality}
  \end{subfigure}

\caption{Three types of visual-linguistic model architectures for cross-modal learning: (a) Cross-modality type, (b) Joint-modality type, and (c) Separate-modality type} \label{fig:vl_model_type}
\end{figure}

Recently, following the success in natural language processing (NLP) tasks~\cite{devlin2019bert}, large-scale BERT-type models have been widely applied to vision-and-language grounding problems~\cite{lu2019vilbert, chen2020uniter, kamath2021mdetr, li2020hero, lei2021less, zellers2021merlot}. The models are usually pre-trained with a large amount of data in both modalities, with standalone tasks that emphasize networks' general ability to understand language and visual information and their interconnection, such as masked language/frame modeling.
When applied to the specific grounding task, the models are fine-tuned with the task's own data to generate certain output, while their ability to reason upon multi-modality input is reserved.
BERT-type models are versatile and powerful,
but considering their scale, the training process is usually time-consuming and computationally challenging.
Moreover, compared to task-specific frameworks, the reasoning flow between the two modalities is less well represented inside BERT.

By observing the existing visual-linguistic models, we categorize their inter-modality reasoning architectures into the following three main types:
1) \textbf{cross-modality} (Fig.~\ref{fig:cross_modality}), such as in ViLBERT~\cite{lu2019vilbert} and LXMERT~\cite{tan2019lxmert}, 
where visual and textual representations are encoded by two separate transformers and then a multi-modal transformer exchanges cross-modal information through a novel co-attentional layer;
2) \textbf{joint-modality} (Fig.~\ref{fig:joint_modality}), such as VL-BERT~\cite{su2019vl} and VideoBERT~\cite{sun2019videobert}, who fuses visual and textual information at the initial stage of a joint transformer model;
3) \textbf{separate-modality} (Fig.~\ref{fig:separate_modality}), such as COOT~\cite{ging2020coot} and T2VLAD~\cite{wang2021t2vlad}, who encodes visual and textual representations with two separate streams, then joints the visual-textual feature space by calculating the inter-modality similarity.
Among these three reasoning types, both cross-modality and joint-modality calculate pairwise attention between $\mathcal{M}$ visual patches and $\mathcal{N}$ textual tokens, requiring $\mathcal{O}(MN)$ time complexity of intra-model information exchange.
On the other hand, separate-modality type models only require a time complexity of $\mathcal{O}(M + N)$.
With better efficiency, it is favored by recent portable video-language applications, and our model falls into this realm.

In this work, we developed a lightweight grounding framework that is versatile for different kinds of video-and-language grounding tasks.
To efficiently train the network, we designed a set of multi-level video-language alignment losses, which are built upon contrastive loss with sophisticated designed positive and negative video-language pairs, to directly shape the feature space.
In the alignment losses, the features of relevant video and language information are aligned respectively in high to low semantic levels and across different time spans.
Using this training scheme, the input from different modalities representing similar subjects is encoded closely in common the feature space,
and this characteristic benefits the downstream video-and-language grounding tasks.
When tested on video QA and retrieval tasks, our method achieved state-of-the-art results on multiple datasets, and it also shows a great potential to be applied in other grounding problems such as video moment retrieval.

\section{Related Work}
\subsection{Vision-and-Language Pre-training}
Driven by revolutionary advances in NLP led by transformer-based pre-training frameworks such as BERT~\cite{devlin2019bert}, GPT2~\cite{radford2019language}, XLNet~\cite{yang2019xlnet}, and GPT3~\cite{brown2020language}, recent years have witnessed a boom in the area of extending the use of transformer architectures to the visual-linguistic tasks. Pioneering works such as ViLBERT~\cite{lu2019vilbert} and LXMERT~\cite{tan2019lxmert} adopt two separate transformers for image and text encoding independently and propose a novel co-attentional transformer layer to fuse visual and linguistic representations. Inspired by the original BERT pre-training tasks, ViLBERT~\cite{lu2019vilbert} is trained through the tasks of reconstruction of the masked image region or text tokens and alignment check if the caption describes the image content. VisualBERT~\cite{li2019visualbert}, Unicoder-VL~\cite{li2020unicoder}, VL-BERT~\cite{su2019vl}, and UNITER~\cite{chen2020uniter} advocate single-stream architecture, where the text and vision sequences are combined as the input of one shared transformer encoder. A new Word-Region Alignment task is proposed by UNITER to further explicitly bridge the fine-grained alignment between visual regions and text tokens. It's worth pointing out that both architectures rely on pre-trained object detectors for extracting ROIs that are viewed as individual visual tokens. A few other works, such as Pixel-BERT~\cite{huang2020pixel} and VirTex~\cite{desai2021virtex} for images or HERO~\cite{li2020hero} for video, operate directly over dense feature maps instead of ROIs extracted by pre-trained object detectors. In these approaches, both visual and textual features are fed into a transformer-based model usually pre-trained with multiple losses. Through these pre-training explorations, tremendous advances have been made in the area of vision-and-language representation learning.

\subsection{Contrastive Learning}
Contrastive learning is a framework that learns such an embedding space that similar sample pairs stay close to each other while dissimilar ones are kept far apart. Contrastive learning can be used in both supervised and unsupervised settings. Remarkable progress has been seen in recent studies in unsupervised visual representation learning~\cite{wu2018unsupervised, van2018representation, ye2019unsupervised, chen2020simple, he2020momentum, chen2020improved} leveraging the power of contrastive learning. We review several popular representative contrastive learning methods that benefit from optimization with negative (dissimilar) samples. Typically, samples in the current mini-batch are utilized in a way that its augmented views are considered as positive samples and are paired with other samples in the same batch as negatives. Computing embeddings for a large number of negative samples in every batch could be computationally prohibitive. As a work-around, memory bank~\cite{wu2018unsupervised} was proposed to store representations of all samples in the dataset from past iterations. The dictionary for each mini-batch is randomly sampled from the memory bank with no back-propagation, so it can support a large dictionary size. However, the representations in the memory bank are from very different encoders all over the past epoch and they are less consistent. MoCo~\cite{he2020momentum} provides a framework of unsupervised learning visual representation as a dynamic dictionary look-up. Compared to memory bank, it enables us to reuse representations of immediately preceding mini-batches due to a queue-based dictionary and is more memory-efficient and can be trained on billion-scale data. SimCLR~\cite{chen2020simple} learns visual representations by maximizing agreement between differently augmented views of the same sample via a contrastive loss in a latent space. It advocates large batch size negatives, stronger data augmentation and introduces the learnable nonlinear transformation, altogether helping improve unsupervised visual representation learning. Our method also benefits from the large-scale negative sample learning. The effects of cross-modal learning without negatives are not discussed in this paper.

\subsection{Video-and-Language Tasks}
Popular video-and-language tasks include text-video retrieval~\cite{chen2011collecting, caba2015activitynet, rohrbach2015dataset, xu2016msr, li2016tgif, wang2019vatex}, video moment retrieval~\cite{zhou2018towards, zhukov2019cross, miech2019howto100m, li2020hero}, video captioning~\cite{chen2011collecting, xu2016msr, krishna2017dense, zhou2018towards, wang2019vatex}, video question answering~\cite{tapaswi2016movieqa, lei2018tvqa, jang2019video, yu2019activityqa, li2020hero, grunde2021agqa}, and video-and-language inference~\cite{liu2020violin}. Text-video retrieval selects a video from a pool of candidate videos, whose content best matches the input text query. Compared to text-image retrieval, text-video retrieval is more challenging that requires the understanding of temporal dynamics and complicated text semantics. Video moment retrieval requires localizing video segments from natural language queries. Video captioning is the task of generating sentences that well describe the input video content, and video question answering aims to predict answers to natural language questions given a video as context. These tasks mainly focus on explicit factual descriptions or explicit information of the video. In contrast, video-and-language inference requires not only explicit visual cues but also more sophisticated reasoning skills, such as inferring reasons and interpreting human emotions. Several efforts~\cite{sun2019videobert, zhu2020actbert, gabeur2020multi, li2020hero, ging2020coot, lei2021less, wang2021t2vlad, liu2021hit} have been made that leverage the powerful transformer architecture as the visual backbones and apply contrastive learning for video-and-language learning tasks. VideoBERT~\cite{sun2019videobert} represents a video with a combined sequence of textual tokens and selected video frames and applies a transformer to learn joint representations. ActBERT~\cite{zhu2020actbert} distinguishes between global actions and local regional objects and encodes them jointly with linguistic descriptions. COOT~\cite{ging2020coot} proposes a hierarchical model that exploits long-range temporal context to produce video-text embedding based on hierarchical interactions between local and global context. Recently, T2VLAD~\cite{wang2021t2vlad} extracts features from the aspects of scene and action, and performs similarity matching with the representations of each local token and the global sentence, while HiT~\cite{liu2021hit} conducts cross-matching between feature-level and semantic-level embedding. But they do not simultaneously decompose the video and text to conduct deep alignment, from where we propose the multi-stream multi-level alignment framework that can be universally applied to various video-and-language tasks.
\section{Methods}
\begin{figure*}[t]
    \centering
    \includegraphics[width=1.\textwidth]{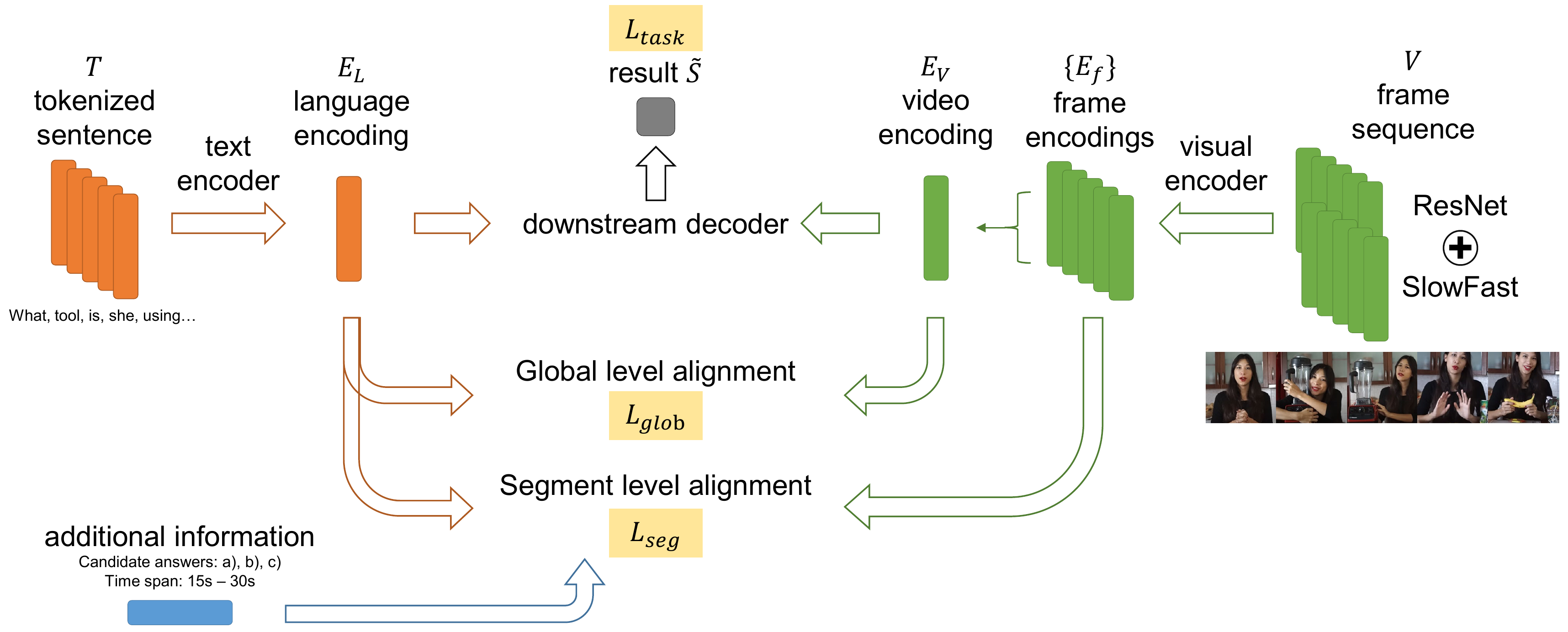}
    \caption{The lightweight video-and-language grounding network and its multi-level alignment training scheme. The video and language inputs are encoded by their individual encoders, then the downstream decoder produces downstream task results. The network is trained with the task-specific loss and multi-level alignment losses.}
    \label{fig:method}
\end{figure*}

In this section, we will discuss the proposed multi-level alignment training scheme in detail.
The training objectives, which models global-level and segment-level alignments, are designed for the network to capture different levels of semantic connection between language and vision.
Together with tasks' specific training loss, the training scheme can be readily applied to various grounding tasks.
The overall framework of our method is described in Figure~\ref{fig:method}.

\subsection{Network Architecture and Notations}
Following the lightweight architecture concept, our network adopts the typical encoder-decoder structure and models inter-modality relation using separate-modality design.
The language and video inputs are encoded with their own encoders, and then passed on to the task-specific decoder to produce downstream answers.

Specifically, our text encoder is implemented as LSTM layers.
It takes tokenized sentence $T$ as the input, and produces every word's encoding $E_{w}$.
After self-attention, words' weighted sum $E_L$ represents the general semantic of the input paragraph.
Meanwhile, the video encoder consists of feature extractors and an extra MLP layers.
We apply two streams of feature extractors, specifically, ResNet~\cite{he2016deep} to extract static object-level visual features and SlowFast~\cite{feichtenhofer2019slowfast} to extract temporal motion-level features,
and the MLP produces each frame's representation $E_f$ from the concatenation of the two.
The video encoding $E_V$ is the weighted sum of every frame's $E_f$, where the weights are computed from a self-attention layer.

For each downstream grounding task, a decoder that generates the task-specific output is designed,
which takes the encodings $E_L$ and $E_V$ and produces the output result $\tilde{S}$.

\subsection{Training overview}
The input of each downstream grounding task is a pair of language and video samples.
These two modalities are semantically related, as their information together infer the result of the task.
Thus, in this work we design two levels of semantic alignment losses, $L_{glob}$ and $L_{seg}$, to enforce the encoders relating the semantic relations between the two modalities, whose details will be discussed in Section~\ref{section:multi_level_alignment}.

For every downstream task, the network will also be trained on the task-specific objective $L_{task}$, e.g., cross-entropy loss for multi-choice video QA task.
As the multi-level alignment losses directly shape the encoders, $L_{task}$ is indispensable to train the decoder in order to obtain reasonable results, further tuning the encoders to learn task-specific information.

To sum up, the overall training loss of our framework is the weighted sum of the objectives mentioned above, where $\lambda_i$ are hyper-parameters to balance individual loss contribution:
\begin{gather}
\label{eq:loss}
    L_{train} = L_{task} + \lambda_1 \cdot L_{glob} + \lambda_2 \cdot L_{seg}
\end{gather}

\begin{figure*}[ht]
\centering

  \begin{subfigure}[c]{0.29\textwidth}
    \includegraphics[width=\linewidth]{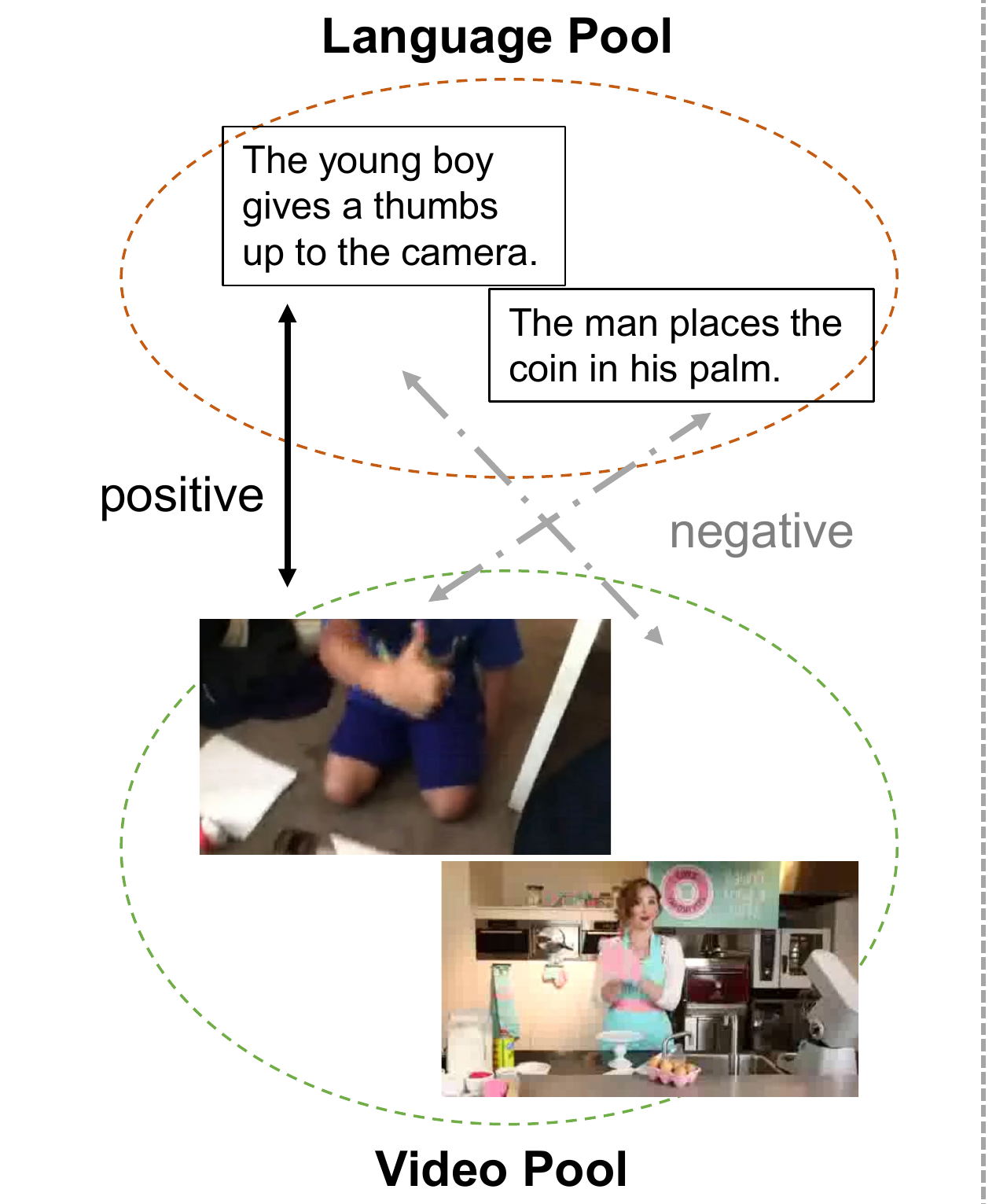}
    \caption{Global Alignment}
    \label{fig:global}
  \end{subfigure}
  \hspace*{\fill}
  \begin{subfigure}[c]{0.69\textwidth}
    \includegraphics[width=\linewidth]{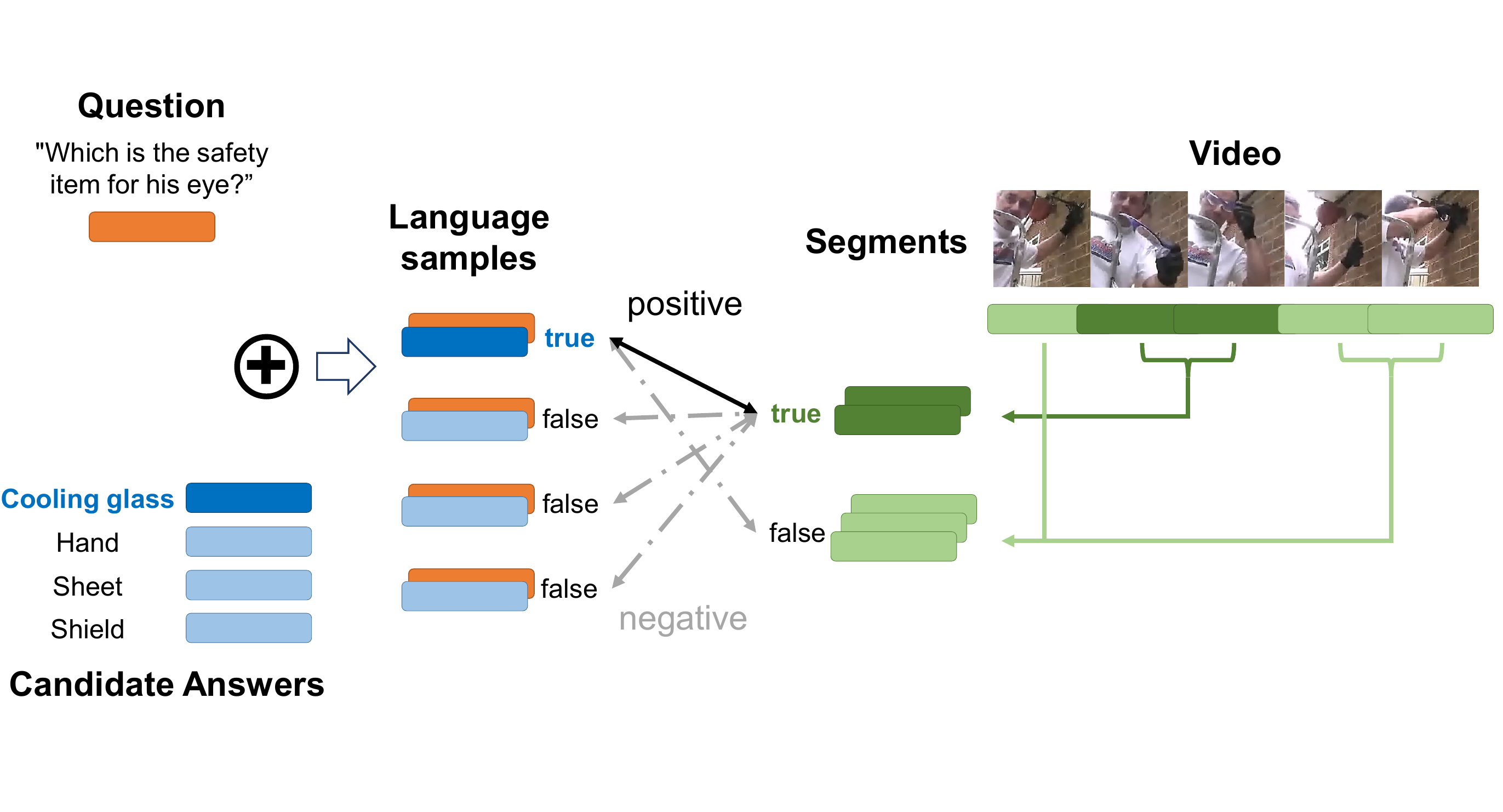}
    \caption{Segment Alignment}
    \label{fig:seg}
  \end{subfigure}

\caption{Multi-level alignment: the positive/negative matching scheme of global level and segment level.} \label{fig:alignments}
\end{figure*}

\subsection{Multi-level Alignment}
\label{section:multi_level_alignment}
Our alignment losses train the network to identify semantic relations by encoding them properly in the feature space.
By contrasting the embedding similarity of modality pairs that are more relevant to the ones that have less semantic connection, the network is able to ground the similar information closer in the shared feature space, bridging the two modalities.
Thus, each of our alignment losses is constructed with the contrastive loss function, where $\alpha$ is a pre-defined margin:
\begin{align}
    L(S^{neg}, S^{pos}) = \left[\alpha +S^{neg} -S^{pos}  \right]_+
\end{align}
It penalizes the similarity score of a negative video-language alignment $S^{neg}$ and encourages a higher similarity score of the positive alignments $S^{pos}$.

The language and video can be decomposed into word-wise and frame-wise information units.
As the units gradually group together, they can convey information from details to overview.
Thus, our embedding matching, i.e., the design of positive/negative pairs in contrastive losses, is conducted on multiple semantic levels.
The global level alignment loss trains the network to capture high-level global information; while the segment level alignment focuses on fine-grained details of language and video inputs.
The positive/negative pairs of these two levels of alignment losses are designed differently; their matching schemes will be discussed in detail below.

\subsubsection{Global Alignment Loss}
The highest level of embedding alignment is conducted globally.
$L_{glob}$ is designed for the network to capture the overall semantic relation of a video clip and its language description, potentially benefiting the video-and-text retrieval task.
It aligns the corresponding video and language pair from large video and language pools.
As shown in Figure~\ref{fig:global}, the language pool contains all the language descriptions of the dataset and the video pool contains all the video clips; 
meanwhile, each description has a video correspondence as the dataset provided.
The embedding similarity of a video-language pair is computed as the cosine distance $\rm cos(*, *)$ between the language and video encoding vectors:
\begin{align}
    S_{glob} = {\rm cos}(E_L, E_{V})
\end{align}

The global alignment loss will contrast the similarity score of the positive video-language pair, which both describe the same theme, to the scores of negative pairs that mismatch.
Since the choice of positive or negative is relevant to the scale of the entire dataset, in practice we use the batch-wise hardest negative alignment in $L_{glob}$:
\begin{align}
    L_{glob} = L(\max_{\{i \in batch\}}{S^{neg_i}_{glob}}, S^{pos}_{glob})
\end{align}
Therefore, with $L_{glob}$, the overall features of language and video are encoded closely in the feature space if they both represent similarly themed information.

\subsubsection{Segment Alignment Loss}
To train the network to capture more fine-grained alignment, we further introduce the segment level alignment loss $L_{seg}$.
For downstream datasets, ground truth sometimes highlights the details and contains additional information that can be used for further supervision.
For instance in video QA tasks, the ground-truth answer can serve as additional knowledge, and the dataset may provide the time-span information on which the question is based. 
As shown in Figure~\ref{fig:seg}, the fine-grained alignment can be constructed using the additional information, benefiting the QA task.
The question combining the correct answer and other candidate answers forms the ``true'' and ``false'' language samples.
Meanwhile, the video clip can also be divided into the ``true'' segment, which contains the frames that the question is based on, and the ``false'' segments that are irrelevant.
The positive alignments are between the ``true'' language sample and the frames from the ``true'' segment, while the negative alignments are between the ``false'' language sample and ``true'' frames or vice versa.
Here, the similarity score of each alignment is computed as
\begin{align}
    S_{seg} = {\rm cos}(\frac{1}{2}(E_L+E_{ans}), E_{f})
\end{align}
where $E_{ans}$ denotes the answer's encoding.

Compared to the global alignment, the segment level alignment focuses on the interior relation of a single video-language context, so the loss $L_{seg}$ is implemented on the data sample-wise:
\begin{align}
    L_{seg} = L(\max_{\{j \in sample\}}{S^{neg_j}_{seg}}, S^{pos}_{seg})
\end{align}
The elaborate matching from $L_{seg}$ further tunes the feature space, and benefits the tasks such as video QA and moment retrieval that generally require understanding subtle details.

\section{Experiment}
Our framework is versatile for various kinds of video-and-language grounding tasks.
We applied our method on two public video QA datasets, How2QA~\cite{li2021value} and ActivityNet-QA~\cite{yu2019activityqa}, and a video-text retrieval dataset, TGIF~\cite{li2016tgif},
where our multi-level alignment training scheme helps the lightweight network achieve comparable results to previous state-of-the-arts.
Besides, we explored the possibility to apply our method to the video moment retrieval task, testing on How2R dataset~\cite{li2021value}.

\subsection{Implementation Details}
For an input video, we used ResNet~\cite{he2016deep} to extract static visual features frame-by-frame and SlowFast~\cite{feichtenhofer2019slowfast} to extract temporal motion features. Both ResNet and SlowFast feature extractors were loaded with pre-trained parameters that were fixed during training. Specifically, we used the implementation of ResNet152 pre-trained on ImageNet data~\cite{deng2009imagenet} from Torchvision and SlowFast pre-trained on Kinetics~\cite{kay2017kinetics}. For every experiment listed in this section, the network was trained for $50$ epochs, with a mini-batch size of $64$. We used AdamW~\cite{loshchilov2017decoupled} to optimize model parameters, with a learning rate $1e-4$, $\beta_1=0.9$ and $\beta_2=0.98$, weight decay $0.01$. Due to the fixed weights of feature extractors, we were able to complete the 50 epochs of either QA or retrieval task training within 1 day on one V100 GPU.

\subsection{Video QA Results}
\begin{table}
\centering
\begin{tabular}{ccc}
\toprule
Dataset & Method & Accuracy (\%) \\
\midrule 
\multirow{2}{*}{\makecell{How2QA \\ \cite{li2021value}}}
    & HERO~\cite{li2020hero} & 60.42 \\
    & Ours & {\bf 63.11} \\
\midrule
\multirow{4}{*}{\makecell{ActivityNet-QA \\ \cite{yu2019activityqa}}}
    & E-SA~\cite{yu2019activityqa} & 31.8 \\
    & CoMVT~\cite{seo2021look} & \underline{36.6} \\
    & VQA-T~\cite{yang2021just} & {\bf 36.8} \\ 
    & Ours & \underline{36.3} \\
\bottomrule
\end{tabular}
\caption{Video QA results: our method compared to the previous state-of-the-arts on the same experimental settings, no subtitle and no pre-training.}
\label{table:QA}
\end{table}

The How2QA dataset~\cite{li2021value} contains Youtube instructional videos with their annotated questions, whose corresponding time spans are also given.
The answer to each question will be chosen from four candidates.
We trained our network on their public training set, which contains $34$k questions, and the result in Table~\ref{table:QA} is reported on the public validation set that contains about $3$k questions.
To get the best result of How2QA dataset, 
we found our best setting is $\lambda_1=0$ and $\lambda_2=1.0$ in the alignment scheme Eq.~\ref{eq:loss}, which means only activating the segment matching constraint.
Although the dataset also provides the subtitle paragraphs of the videos, we did not include this information in training and testing; 
when compared to the previous state-of-the-art method HERO~\cite{li2020hero} in the same setting (without pre-training), our lightweight model achieved better performance than the BERT-type model, with an improvement in accuracy of absolute $2.69\%$.

The ActivityNet-QA dataset~\cite{yu2019activityqa} contains $58$k QA pairs that come from $5.8$k activity videos.
The questions are open-ended, but we took the most frequently occurred $1$k answers as the candidates.
In our segment-level alignment, we randomly picked three ``false'' answers from the candidates, and as no time span information was provided, no ``false'' video segment was contrasted.
Meanwhile, in the global-level matching, the language pool was the set of question-answer pairs in the batch while the video pool contained the videos.
When trained with $\lambda_1=1.0$ and $\lambda_2=2.0$, our network achieved $36.3\%$ accuracy on the public test set, which is comparable to the performance of previous state-of-the-art methods~\cite{seo2021look,yang2021just} in the same non-pre-training setting.

\begin{table}
\centering
\begin{tabular}{ccc}
\toprule
Study & Method & Accuracy (\%) \\
\midrule 
\multirow{4}{*}{Alignments}
    & None & 59.84 \\
    & Global & 60.90 \\
    & Segment & {\bf 63.11} \\
    & Global + Segment & 61.96 \\
\midrule
\multirow{3}{*}{\makecell{Feature \\ extractors}}
    & ResNet & 62.47 \\
    & SlowFast & 62.63 \\
    & ResNet + SlowFast & {\bf 63.11} \\
\bottomrule
\end{tabular}
\caption{Ablation studies: the effect of different alignment loss combinations and feature extractors, tested on How2QA~\cite{li2021value}.}
\label{table:ablation}
\end{table}

\subsubsection{Ablation Study}
\paragraph{Multi-level alignments}
    To study the effect of our multi-level alignment scheme, on the How2QA dataset we did the ablation study of different loss combinations. 
    As shown in the first section of Table~\ref{table:ablation}, when the network is only trained with task-specific loss and no alignment loss was applied, the accuracy is lower than $60\%$.
    When one of the alignment losses was added, the performance improved, and the best result came from adding the segment level matching.
    However, we notice that when both levels of alignment were applied, the performance was slightly worse than applying the segment alignment alone.
    The possible reason may be the characteristic of this dataset, where the question can only correspond to a small fraction of the video, 
    thus the global alignment may cause confusion with information from irrelevant frames while the network simultaneously learning the more fine-grained segment alignment.
    This finding informs future applications to activate alignment levels differently according to the feature of their data.
\paragraph{Visual features}
    As we applied two streams of feature extractors in our framework (Fig.~\ref{fig:method}), we also studied the necessity of using them both.
    As shown in the second section of Table~\ref{table:ablation}, when the network was only given one type of visual features, i.e., ResNet static features or SlowFast motion feature, the performance decreased slightly, indicating the merit of applying both types of features.

\begin{figure*}[ht]
    \centering
    \includegraphics[width=1.\textwidth]{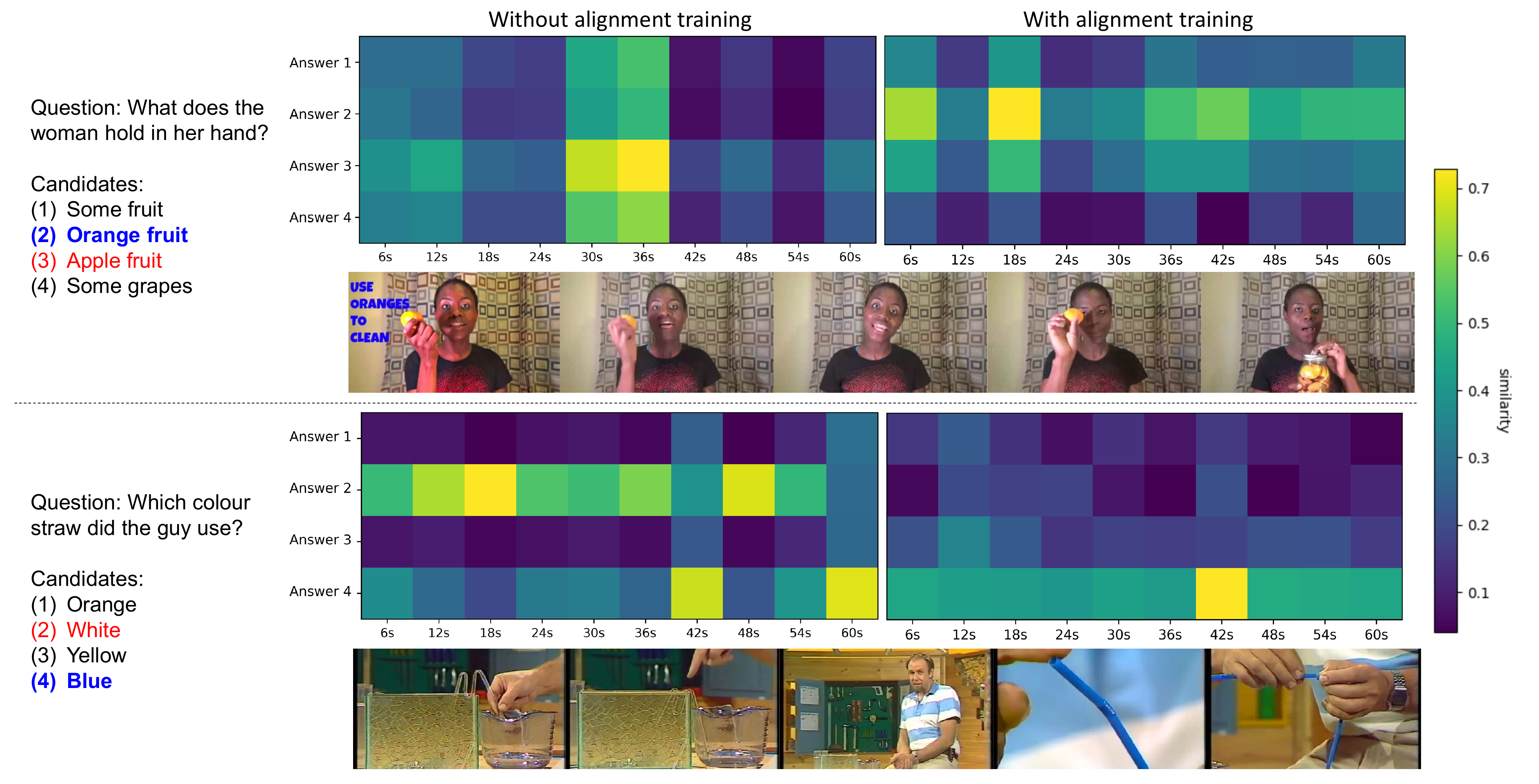}
    \caption{The heatmaps show the encoding feature vectors' similarity between each pair of question+candidate and video frame. When the network was not supervised with alignment losses, the correlation between two modalities was less clear; \textcolor{red}{wrong answers} (marked as red) may get more ``attention'' at the wrong time, as the heatmaps shows on the left. When the alignment loss was added, the predicted correlation became more reasonable, leading to the \textcolor{blue}{correct predictions} (marked as blue).}
    \label{fig:visual}
\end{figure*}

\subsubsection{Alignment Visualization}
To further analyze the effect of alignment losses, we visualize the correlation between the encodings of two modalities.
In Fig.~\ref{fig:visual}, we show two cases from the How2QA test set, with the results from the network trained without or with alignment losses.
The heatmaps show the cosine similarity between the encoding features of each language (question+candidate answer) and frame pair.
The brighter the color is, the stronger the vision and language are correlated to each other.

In the first example, the person is showing an orange to the camera at the beginning of the video and near the end.
For the network trained without the alignment scheme, it failed to correlate the ``holding'' action with the corresponding frames of the video, predicting the wrong answer.
However, the network trained with our alignments successfully recognized the action in early frames as the similarities there are higher, therefore giving the right answer.
Similarly, in the second case, the feature similarity between the correct answer (blue straw) and the frames in the corresponding timespan is the highest when alignment losses were used.
On the other hand, without the alignment training, the information from irrelevant frames distracted the network, causing it to predict the straw as white.

The visualization of the feature similarities in these test cases clearly shows the benefit of using additional alignment information in the training.
The correlation between the two modalities is better learned by the network, leading to better performance on the QA task.

\subsection{Video-text Retrieval Results}

\begin{table}[ht]
\small
\centering
\begin{tabular}{ccc}
\toprule
    \multirow{2}{*}{Method} & \multicolumn{2}{c}{Retrieval R@1} \\
    \cmidrule{2-3}
    & Text-to-video & Video-to-text \\
\midrule
    DeViSE~\cite{frome2013devise} & 2.2 & 2.1\\
    Corr-AE~\cite{feng2014cross} & 2.1 & 2.2 \\
    Order~\cite{vendrov2015order} & 1.6 & 1.7 \\
    VSE++~\cite{faghri2017vse++} & 1.6 & 1.4 \\ 
    PVSE~\cite{song2019polysemous} & 3.0 & 3.3 \\ 
    HGR~\cite{chen2020fine} & 4.5 & -- \\
    Ours & {\bf 4.7} & {\bf 6.1} \\
\bottomrule
\end{tabular}
\caption{Video-text retrieval results on TGIF dataset~\cite{li2016tgif}. The results of previous works were reported in literatures~\cite{song2019polysemous,chen2020fine}.}
\label{table:retrieval}
\end{table}

Our framework can also be applied to the video-text retrieval task, where the network's job is to find the corresponding text/video of the given video/text.
In this application, global level alignment is activated, which alone should be capable for the job; meanwhile, the task-specific loss function is not needed and segment matching is not applicable.
We tested our method on the TGIF dataset~\cite{li2016tgif}, which contains short gif videos and their descriptions. 
We used their training set that contains $80$k videos for the training, and the results in Table~\ref{table:retrieval} are reported on the official test set with $11$k videos.
Compared to the previous methods~\cite{song2019polysemous,chen2020fine} that were specifically designed for retrieval, our framework can achieve the state-of-the-art result of the recall at the top $1$ ranking, 
showing the capability of our method to be applied to different types of grounding tasks.

\subsection{Video Moment Retrieval Results}

\begin{table}[ht]
\small
\centering
\begin{tabular}{ccc}
\toprule
Alignment & tIoU $\ge0.5$ (\%) & tIoU $\ge0.7$ (\%) \\
\midrule 
None & 9.2 & 2.5 \\
Global & 10.9 & {\bf 3.9} \\
Segment & 10.4 & 3.0 \\
Global + Segment & {\bf 11.0} & 3.8 \\
\bottomrule
\end{tabular}
\caption{Our approach's video moment retrieval results on How2R dataset~\cite{li2021value}, under different alignment settings.}
\label{table:localization}
\end{table}

Lastly, we explored the possibility to apply our multi-level alignments on the video moment retrieval task How2R~\cite{li2021value}, where the description of a certain segment of the video is given, and the network needs to predict the time span of that segment.
For our global level alignment, language and video pools were implemented as the collections of the data in the mini-batch; in the segment level, no ``false'' language sample was contrasted.
We used the cross-entropy loss as the task-specific loss for the network to further learn the correct starting and ending timestamps. 

Table~\ref{table:localization} lists our approach's results under different alignment settings, evaluating under temporal Intersection over Union (tIoU) that measures the overlap between the predicted span and the ground-truth span.
Compared to the network trained without any alignment loss, the performance improved as we added either global or segment alignment loss to further constrain the training.
This indicates the merit of alignment losses in video moment retrieval tasks.
However, we notice that using both alignments did not outperform applying only one alignment.
This might due to the nature of the dataset, and better hyper-parameter searching to balance global-segment information might potentially improve the performance.
\section{Conclusion}
In this work, we developed a multi-level alignment training scheme for video-and-language grounding tasks.
To better encode the visual and linguistic modalities in the shared feature space, global level alignment loss focuses on training the network to capture context information, while the segment level alignment emphasizes fine-grained semantics.
The application scenario of our multi-level alignment scheme is not restricted.
It can be applied to video QA, video-text retrieval and video moment retrieval tasks, and was able to help the lightweight model achieve good performance on multiple datasets.

\bibliographystyle{IEEEtran}
\bibliography{ref}

\begin{thebibliography}{10}
\providecommand{\url}[1]{#1}
\csname url@samestyle\endcsname
\providecommand{\newblock}{\relax}
\providecommand{\bibinfo}[2]{#2}
\providecommand{\BIBentrySTDinterwordspacing}{\spaceskip=0pt\relax}
\providecommand{\BIBentryALTinterwordstretchfactor}{4}
\providecommand{\BIBentryALTinterwordspacing}{\spaceskip=\fontdimen2\font plus
\BIBentryALTinterwordstretchfactor\fontdimen3\font minus
  \fontdimen4\font\relax}
\providecommand{\BIBforeignlanguage}[2]{{%
\expandafter\ifx\csname l@#1\endcsname\relax
\typeout{** WARNING: IEEEtran.bst: No hyphenation pattern has been}%
\typeout{** loaded for the language `#1'. Using the pattern for}%
\typeout{** the default language instead.}%
\else
\language=\csname l@#1\endcsname
\fi
#2}}
\providecommand{\BIBdecl}{\relax}
\BIBdecl

\bibitem{xu2019scene}
N.~Xu, A.-A. Liu, J.~Liu, W.~Nie, and Y.~Su, ``Scene graph captioner: Image
  captioning based on structural visual representation,'' \emph{Journal of
  Visual Communication and Image Representation}, vol.~58, pp. 477--485, 2019.

\bibitem{lee2019visual}
S.~Lee, J.-W. Kim, Y.~Oh, and J.~H. Jeon, ``Visual question answering over
  scene graph,'' in \emph{2019 First International Conference on Graph
  Computing (GC)}.\hskip 1em plus 0.5em minus 0.4em\relax IEEE, 2019, pp.
  45--50.

\bibitem{liang2020lrta}
W.~Liang, F.~Niu, A.~Reganti, G.~Thattai, and G.~Tur, ``Lrta: A transparent
  neural-symbolic reasoning framework with modular supervision for visual
  question answering,'' \emph{arXiv preprint arXiv:2011.10731}, 2020.

\bibitem{chen2020uniter}
Y.-C. Chen, L.~Li, L.~Yu, A.~El~Kholy, F.~Ahmed, Z.~Gan, Y.~Cheng, and J.~Liu,
  ``Uniter: Universal image-text representation learning,'' in \emph{European
  conference on computer vision}.\hskip 1em plus 0.5em minus 0.4em\relax
  Springer, 2020, pp. 104--120.

\bibitem{zhu2020actbert}
L.~Zhu and Y.~Yang, ``Actbert: Learning global-local video-text
  representations,'' in \emph{Proceedings of the IEEE/CVF conference on
  computer vision and pattern recognition}, 2020, pp. 8746--8755.

\bibitem{chen2020fine}
S.~Chen, Y.~Zhao, Q.~Jin, and Q.~Wu, ``Fine-grained video-text retrieval with
  hierarchical graph reasoning,'' \emph{CVPR}, 2020.

\bibitem{le2020hierarchical}
T.~M. Le, V.~Le, S.~Venkatesh, and T.~Tran, ``Hierarchical conditional relation
  networks for video question answering,'' in \emph{Proceedings of the IEEE/CVF
  conference on computer vision and pattern recognition}, 2020, pp. 9972--9981.

\bibitem{devlin2019bert}
J.~Devlin, M.-W. Chang, K.~Lee, and K.~Toutanova, ``Bert: Pre-training of deep
  bidirectional transformers for language understanding,'' in \emph{NAACL-HLT
  (1)}, 2019.

\bibitem{lu2019vilbert}
J.~Lu, D.~Batra, D.~Parikh, and S.~Lee, ``Vilbert: Pretraining task-agnostic
  visiolinguistic representations for vision-and-language tasks,''
  \emph{Advances in neural information processing systems}, vol.~32, 2019.

\bibitem{kamath2021mdetr}
A.~Kamath, M.~Singh, Y.~LeCun, G.~Synnaeve, I.~Misra, and N.~Carion,
  ``Mdetr-modulated detection for end-to-end multi-modal understanding,'' in
  \emph{Proceedings of the IEEE/CVF International Conference on Computer
  Vision}, 2021, pp. 1780--1790.

\bibitem{li2020hero}
L.~Li, Y.-C. Chen, Y.~Cheng, Z.~Gan, L.~Yu, and J.~Liu, ``Hero: Hierarchical
  encoder for video+ language omni-representation pre-training,'' in
  \emph{EMNLP}, 2020.

\bibitem{lei2021less}
J.~Lei, L.~Li, L.~Zhou, Z.~Gan, T.~L. Berg, M.~Bansal, and J.~Liu, ``Less is
  more: Clipbert for video-and-language learning via sparse sampling,'' in
  \emph{Proceedings of the IEEE/CVF Conference on Computer Vision and Pattern
  Recognition}, 2021, pp. 7331--7341.

\bibitem{zellers2021merlot}
R.~Zellers, X.~Lu, J.~Hessel, Y.~Yu, J.~S. Park, J.~Cao, A.~Farhadi, and
  Y.~Choi, ``Merlot: Multimodal neural script knowledge models,''
  \emph{Advances in Neural Information Processing Systems}, vol.~34, 2021.

\bibitem{tan2019lxmert}
H.~Tan and M.~Bansal, ``Lxmert: Learning cross-modality encoder representations
  from transformers,'' in \emph{Proceedings of the 2019 Conference on Empirical
  Methods in Natural Language Processing}, 2019.

\bibitem{su2019vl}
\BIBentryALTinterwordspacing
W.~Su, X.~Zhu, Y.~Cao, B.~Li, L.~Lu, F.~Wei, and J.~Dai, ``Vl-bert:
  Pre-training of generic visual-linguistic representations,'' in
  \emph{International Conference on Learning Representations}, 2020. [Online].
  Available: \url{https://openreview.net/forum?id=SygXPaEYvH}
\BIBentrySTDinterwordspacing

\bibitem{sun2019videobert}
C.~Sun, A.~Myers, C.~Vondrick, K.~Murphy, and C.~Schmid, ``Videobert: A joint
  model for video and language representation learning,'' in \emph{Proceedings
  of the IEEE/CVF International Conference on Computer Vision}, 2019, pp.
  7464--7473.

\bibitem{ging2020coot}
S.~Ging, M.~Zolfaghari, H.~Pirsiavash, and T.~Brox, ``Coot: Cooperative
  hierarchical transformer for video-text representation learning,''
  \emph{Advances in neural information processing systems}, vol.~33, pp.
  22\,605--22\,618, 2020.

\bibitem{wang2021t2vlad}
X.~Wang, L.~Zhu, and Y.~Yang, ``T2vlad: global-local sequence alignment for
  text-video retrieval,'' in \emph{Proceedings of the IEEE/CVF Conference on
  Computer Vision and Pattern Recognition}, 2021, pp. 5079--5088.

\bibitem{radford2019language}
A.~Radford, J.~Wu, R.~Child, D.~Luan, D.~Amodei, I.~Sutskever \emph{et~al.},
  ``Language models are unsupervised multitask learners,'' \emph{OpenAI blog},
  vol.~1, no.~8, p.~9, 2019.

\bibitem{yang2019xlnet}
Z.~Yang, Z.~Dai, Y.~Yang, J.~Carbonell, R.~R. Salakhutdinov, and Q.~V. Le,
  ``Xlnet: Generalized autoregressive pretraining for language understanding,''
  \emph{Advances in neural information processing systems}, vol.~32, 2019.

\bibitem{brown2020language}
T.~Brown, B.~Mann, N.~Ryder, M.~Subbiah, J.~D. Kaplan, P.~Dhariwal,
  A.~Neelakantan, P.~Shyam, G.~Sastry, A.~Askell \emph{et~al.}, ``Language
  models are few-shot learners,'' \emph{Advances in neural information
  processing systems}, vol.~33, pp. 1877--1901, 2020.

\bibitem{li2019visualbert}
L.~H. Li, M.~Yatskar, D.~Yin, C.-J. Hsieh, and K.-W. Chang, ``Visualbert: A
  simple and performant baseline for vision and language,'' \emph{arXiv
  preprint arXiv:1908.03557}, 2019.

\bibitem{li2020unicoder}
G.~Li, N.~Duan, Y.~Fang, M.~Gong, and D.~Jiang, ``Unicoder-vl: A universal
  encoder for vision and language by cross-modal pre-training,'' in
  \emph{Proceedings of the AAAI Conference on Artificial Intelligence},
  vol.~34, no.~07, 2020, pp. 11\,336--11\,344.

\bibitem{huang2020pixel}
Z.~Huang, Z.~Zeng, B.~Liu, D.~Fu, and J.~Fu, ``Pixel-bert: Aligning image
  pixels with text by deep multi-modal transformers,'' \emph{arXiv preprint
  arXiv:2004.00849}, 2020.

\bibitem{desai2021virtex}
K.~Desai and J.~Johnson, ``Virtex: Learning visual representations from textual
  annotations,'' in \emph{Proceedings of the IEEE/CVF Conference on Computer
  Vision and Pattern Recognition}, 2021, pp. 11\,162--11\,173.

\bibitem{wu2018unsupervised}
Z.~Wu, Y.~Xiong, S.~X. Yu, and D.~Lin, ``Unsupervised feature learning via
  non-parametric instance discrimination,'' in \emph{Proceedings of the IEEE
  conference on computer vision and pattern recognition}, 2018, pp. 3733--3742.

\bibitem{van2018representation}
A.~Van~den Oord, Y.~Li, and O.~Vinyals, ``Representation learning with
  contrastive predictive coding,'' \emph{arXiv e-prints}, pp. arXiv--1807,
  2018.

\bibitem{ye2019unsupervised}
M.~Ye, X.~Zhang, P.~C. Yuen, and S.-F. Chang, ``Unsupervised embedding learning
  via invariant and spreading instance feature,'' in \emph{Proceedings of the
  IEEE/CVF Conference on Computer Vision and Pattern Recognition}, 2019, pp.
  6210--6219.

\bibitem{chen2020simple}
T.~Chen, S.~Kornblith, M.~Norouzi, and G.~Hinton, ``A simple framework for
  contrastive learning of visual representations,'' in \emph{International
  conference on machine learning}.\hskip 1em plus 0.5em minus 0.4em\relax PMLR,
  2020, pp. 1597--1607.

\bibitem{he2020momentum}
K.~He, H.~Fan, Y.~Wu, S.~Xie, and R.~Girshick, ``Momentum contrast for
  unsupervised visual representation learning,'' in \emph{Proceedings of the
  IEEE/CVF conference on computer vision and pattern recognition}, 2020, pp.
  9729--9738.

\bibitem{chen2020improved}
X.~Chen, H.~Fan, R.~Girshick, and K.~He, ``Improved baselines with momentum
  contrastive learning,'' \emph{arXiv preprint arXiv:2003.04297}, 2020.

\bibitem{chen2011collecting}
D.~Chen and W.~B. Dolan, ``Collecting highly parallel data for paraphrase
  evaluation,'' in \emph{Proceedings of the 49th annual meeting of the
  association for computational linguistics: human language technologies},
  2011, pp. 190--200.

\bibitem{caba2015activitynet}
F.~Caba~Heilbron, V.~Escorcia, B.~Ghanem, and J.~Carlos~Niebles, ``Activitynet:
  A large-scale video benchmark for human activity understanding,'' in
  \emph{Proceedings of the ieee conference on computer vision and pattern
  recognition}, 2015, pp. 961--970.

\bibitem{rohrbach2015dataset}
A.~Rohrbach, M.~Rohrbach, N.~Tandon, and B.~Schiele, ``A dataset for movie
  description,'' in \emph{Proceedings of the IEEE conference on computer vision
  and pattern recognition}, 2015, pp. 3202--3212.

\bibitem{xu2016msr}
J.~Xu, T.~Mei, T.~Yao, and Y.~Rui, ``Msr-vtt: A large video description dataset
  for bridging video and language,'' in \emph{Proceedings of the IEEE
  conference on computer vision and pattern recognition}, 2016, pp. 5288--5296.

\bibitem{li2016tgif}
Y.~Li, Y.~Song, L.~Cao, J.~Tetreault, L.~Goldberg, A.~Jaimes, and J.~Luo,
  ``Tgif: A new dataset and benchmark on animated gif description,'' in
  \emph{Proceedings of the IEEE Conference on Computer Vision and Pattern
  Recognition}, 2016, pp. 4641--4650.

\bibitem{wang2019vatex}
X.~Wang, J.~Wu, J.~Chen, L.~Li, Y.-F. Wang, and W.~Y. Wang, ``Vatex: A
  large-scale, high-quality multilingual dataset for video-and-language
  research,'' in \emph{Proceedings of the IEEE/CVF International Conference on
  Computer Vision}, 2019, pp. 4581--4591.

\bibitem{zhou2018towards}
L.~Zhou, C.~Xu, and J.~J. Corso, ``Towards automatic learning of procedures
  from web instructional videos,'' in \emph{Thirty-Second AAAI Conference on
  Artificial Intelligence}, 2018.

\bibitem{zhukov2019cross}
D.~Zhukov, J.-B. Alayrac, R.~G. Cinbis, D.~Fouhey, I.~Laptev, and J.~Sivic,
  ``Cross-task weakly supervised learning from instructional videos,'' in
  \emph{Proceedings of the IEEE/CVF Conference on Computer Vision and Pattern
  Recognition}, 2019, pp. 3537--3545.

\bibitem{miech2019howto100m}
A.~Miech, D.~Zhukov, J.-B. Alayrac, M.~Tapaswi, I.~Laptev, and J.~Sivic,
  ``Howto100m: Learning a text-video embedding by watching hundred million
  narrated video clips,'' in \emph{Proceedings of the IEEE/CVF International
  Conference on Computer Vision}, 2019, pp. 2630--2640.

\bibitem{krishna2017dense}
R.~Krishna, K.~Hata, F.~Ren, L.~Fei-Fei, and J.~Carlos~Niebles,
  ``Dense-captioning events in videos,'' in \emph{Proceedings of the IEEE
  international conference on computer vision}, 2017, pp. 706--715.

\bibitem{tapaswi2016movieqa}
M.~Tapaswi, Y.~Zhu, R.~Stiefelhagen, A.~Torralba, R.~Urtasun, and S.~Fidler,
  ``Movieqa: Understanding stories in movies through question-answering,'' in
  \emph{Proceedings of the IEEE conference on computer vision and pattern
  recognition}, 2016, pp. 4631--4640.

\bibitem{lei2018tvqa}
J.~Lei, L.~Yu, M.~Bansal, and T.~Berg, ``Tvqa: Localized, compositional video
  question answering,'' in \emph{Proceedings of the 2018 Conference on
  Empirical Methods in Natural Language Processing}, 2018, pp. 1369--1379.

\bibitem{jang2019video}
Y.~Jang, Y.~Song, C.~D. Kim, Y.~Yu, Y.~Kim, and G.~Kim, ``Video question
  answering with spatio-temporal reasoning,'' \emph{International Journal of
  Computer Vision}, vol. 127, no.~10, pp. 1385--1412, 2019.

\bibitem{yu2019activityqa}
Z.~Yu, D.~Xu, J.~Yu, T.~Yu, Z.~Zhao, Y.~Zhuang, and D.~Tao, ``Activitynet-qa: A
  dataset for understanding complex web videos via question answering,'' in
  \emph{AAAI}, 2019, pp. 9127--9134.

\bibitem{grunde2021agqa}
M.~Grunde-McLaughlin, R.~Krishna, and M.~Agrawala, ``Agqa: A benchmark for
  compositional spatio-temporal reasoning,'' in \emph{Proceedings of the
  IEEE/CVF Conference on Computer Vision and Pattern Recognition}, 2021, pp.
  11\,287--11\,297.

\bibitem{liu2020violin}
J.~Liu, W.~Chen, Y.~Cheng, Z.~Gan, L.~Yu, Y.~Yang, and J.~Liu, ``Violin: A
  large-scale dataset for video-and-language inference,'' in \emph{Proceedings
  of the IEEE/CVF Conference on Computer Vision and Pattern Recognition}, 2020,
  pp. 10\,900--10\,910.

\bibitem{gabeur2020multi}
V.~Gabeur, C.~Sun, K.~Alahari, and C.~Schmid, ``Multi-modal transformer for
  video retrieval,'' in \emph{European Conference on Computer Vision}.\hskip
  1em plus 0.5em minus 0.4em\relax Springer, 2020, pp. 214--229.

\bibitem{liu2021hit}
S.~Liu, H.~Fan, S.~Qian, Y.~Chen, W.~Ding, and Z.~Wang, ``Hit: Hierarchical
  transformer with momentum contrast for video-text retrieval,'' in
  \emph{Proceedings of the IEEE/CVF International Conference on Computer
  Vision}, 2021, pp. 11\,915--11\,925.

\bibitem{he2016deep}
K.~He, X.~Zhang, S.~Ren, and J.~Sun, ``Deep residual learning for image
  recognition,'' in \emph{Proceedings of the IEEE conference on computer vision
  and pattern recognition}, 2016, pp. 770--778.

\bibitem{feichtenhofer2019slowfast}
C.~Feichtenhofer, H.~Fan, J.~Malik, and K.~He, ``Slowfast networks for video
  recognition,'' in \emph{Proceedings of the IEEE/CVF international conference
  on computer vision}, 2019, pp. 6202--6211.

\bibitem{li2021value}
L.~Li, J.~Lei, Z.~Gan, L.~Yu, Y.-C. Chen, R.~Pillai, Y.~Cheng, L.~Zhou, X.~E.
  Wang, W.~Y. Wang \emph{et~al.}, ``Value: A multi-task benchmark for
  video-and-language understanding evaluation,'' in \emph{35th Conference on
  Neural Information Processing Systems (NeurIPS 2021) Track on Datasets and
  Benchmarks}, 2021.

\bibitem{deng2009imagenet}
J.~Deng, W.~Dong, R.~Socher, L.-J. Li, K.~Li, and L.~Fei-Fei, ``Imagenet: A
  large-scale hierarchical image database,'' in \emph{2009 IEEE conference on
  computer vision and pattern recognition}.\hskip 1em plus 0.5em minus
  0.4em\relax Ieee, 2009, pp. 248--255.

\bibitem{kay2017kinetics}
W.~Kay, J.~Carreira, K.~Simonyan, B.~Zhang, C.~Hillier, S.~Vijayanarasimhan,
  F.~Viola, T.~Green, T.~Back, P.~Natsev \emph{et~al.}, ``The kinetics human
  action video dataset,'' \emph{arXiv preprint arXiv:1705.06950}, 2017.

\bibitem{loshchilov2017decoupled}
I.~Loshchilov and F.~Hutter, ``Decoupled weight decay regularization,''
  \emph{arXiv preprint arXiv:1711.05101}, 2017.

\bibitem{seo2021look}
P.~H. Seo, A.~Nagrani, and C.~Schmid, ``Look before you speak: Visually
  contextualized utterances,'' in \emph{Proceedings of the IEEE/CVF Conference
  on Computer Vision and Pattern Recognition}, 2021, pp. 16\,877--16\,887.

\bibitem{yang2021just}
A.~Yang, A.~Miech, J.~Sivic, I.~Laptev, and C.~Schmid, ``Just ask: Learning to
  answer questions from millions of narrated videos,'' in \emph{Proceedings of
  the IEEE/CVF International Conference on Computer Vision}, 2021, pp.
  1686--1697.

\bibitem{frome2013devise}
A.~Frome, G.~S. Corrado, J.~Shlens, S.~Bengio, J.~Dean, M.~Ranzato, and
  T.~Mikolov, ``Devise: A deep visual-semantic embedding model,''
  \emph{Advances in neural information processing systems}, vol.~26, 2013.

\bibitem{feng2014cross}
F.~Feng, X.~Wang, and R.~Li, ``Cross-modal retrieval with correspondence
  autoencoder,'' in \emph{Proceedings of the 22nd ACM international conference
  on Multimedia}, 2014, pp. 7--16.

\bibitem{vendrov2015order}
I.~Vendrov, R.~Kiros, S.~Fidler, and R.~Urtasun, ``Order-embeddings of images
  and language,'' \emph{arXiv preprint arXiv:1511.06361}, 2015.

\bibitem{faghri2017vse++}
F.~Faghri, D.~J. Fleet, J.~R. Kiros, and S.~Fidler, ``Vse++: Improving
  visual-semantic embeddings with hard negatives,'' \emph{arXiv preprint
  arXiv:1707.05612}, 2017.

\bibitem{song2019polysemous}
Y.~Song and M.~Soleymani, ``Polysemous visual-semantic embedding for
  cross-modal retrieval,'' in \emph{Proceedings of the IEEE/CVF Conference on
  Computer Vision and Pattern Recognition}, 2019, pp. 1979--1988.

\end{thebibliography}

\end{document}